# Few-shot Object Detection with Self-adaptive Attention Network for Remote Sensing Images

Zixuan Xiao, Wei Xue, and Ping Zhong, *Senior Member, IEEE*

*Abstract*—In remote sensing field, there are many applications of object detection in recent years, which demands a great number of labeled data. However, we may be faced with some cases where only limited data are available. In this paper, we proposed a few-shot object detector which is designed for detecting novel objects provided with only a few examples. Particularly, in order to fit the object detection settings, our proposed few-shot detector concentrates on the relations that lie in the level of objects instead of the full image with the assistance of Self-Adaptive Attention Network (SAAN). The SAAN can fully leverage the object-level relations through a relation GRU unit and simultaneously attach attention on object features in a self-adaptive way according to the object-level relations to avoid some situations where the additional attention is useless or even detrimental. Eventually, the detection results are produced from the features that are added with attention and thus are able to be detected simply. The experiments demonstrate the effectiveness of the proposed method in few-shot scenes.

## I. INTRODUCTION

THANKS to the development recently in computer vision, the past few decades have seen the rapid progress in remote sensing technology which has brought quantities of applications [1]-[4], such as environmental management, forecasts of disasters and assistance in rescue operations. Development of VHR (very high resolution) remote sensing images provide us with more detailed geo-spatial objects information including diversities in scale, orientation, and shape. Among these applications, object detection, which is one of the main tasks in remote sensing, has played an important part with the assistance of the deep convolutional neural networks (CNNs). Whereas as we all know, methods based on deep CNNs rely heavily on a large number of training data. However, in some cases, the training data are not easy to obtain. Therefore, the training data are rare in these cases, which may lead to poor performance as deep CNNs severely overfit and fail to generalize. This bottleneck may be even worse for the complex object detection task.

The research targeting at the condition where the training data are extremely rare is called few-shot learning (FSL) [5]. In FSL, there are many works concentrating on classification task in natural images. However, in this paper, we focus on the challenging task, i.e. few-shot object detection in remote sensing. In detail, we aim to make the model able to detect the novel objects when there are some base classes with sufficient number of annotations and novel classes with only a few annotations (1~10 shots). In FSL field, the research [6]-[9] mainly focuses on the classification task that is much easier than object detection task. In fact, the difficulty comes from that the object detection task should not only offer us which class the object belongs to like what the classification task does, but also tell us where it is in the image. On the other hand, when we face few-shot cases in classification task, a few images of the class are available and thus the data lie in image-level. Whereas in the detection task, only a few annotations that lie in object-level are available, which is quite different from the former. In other words, the data in classification task mainly lie in image-level whereas the one in object detection task lies in object-level. Moreover, with multiple objects in a single image, detecting object in novel classes may be interfered by objects from other classes to a large extent, especially for VHR remote sensing images. As a consequence, it is not feasible to apply the methods that aim to handle few-shot classification task on the few-shot object detection task.

Therefore, it is necessary to concentrate on object-level features. Our proposed model is based on the RoI (Region-of-Interest) features from two-stage detector, such as Faster R-CNN [10] in an intuitive but effective way. In our model, there are mainly two networks: The first one is an object detection network the same as common detection network. The other one is our Self-Adaptive Attention Network (SAAN). Particularly, the parameters of the detection network are pre-trained and the ones of the backbone in our SAAN is shared with the first part to avoid additional parameters. All of these settings are designed to fit the few-shot situation. The same as the way some meta-learning and few-shot detection works do to process training images, we define support set as well as query set. In query set, the images are definitely the one from the detection dataset. Whereas in support set, the images are parts of objects cropped from the same detection dataset in a simply designed way to be introduced in detail subsequently. In a single training

This work was supported in part by the National Science Foundation of China under Grant 61671456 and Grant 61271439, in part by the Foundation for the Author of National Excellent Doctoral Dissertation of China under Grant 201243, and in part by the Program for New Century Excellent Talents in University under Grant NECT-13-0164. *(Corresponding author: Ping Zhong.)*

Z. Xiao, W. Xue, and P. Zhong are with the National Key Laboratory of Science and Technology on Automatic Target Recognition, College of Electrical Science and Technology, National University of Defense Technology, Changsha, 410073, China; (e-mail: xiaozixuan18@nudt.edu.cn, cswxue@ahut.edu.cn, zhongping@nudt.edu.cn).

W. Xue is also with the School of Computer Science and Technology, Anhui University of Technology, Maanshan 243032, China.

step, the inputs are a combination of a query image and some support images whose number is equal to the number of classes to be detect. In detail, a query image is taken by the object detection network as input to obtain RoI features. Simultaneously, the support images are fed into the SAAN. After that, the so-called attention features which contain the information from the few-shot objects are produced. In a self-adaptive way, these features are implemented to be fused into the few-shot object features. Eventually, the self-adaptively attention-added features are fed into the predictor-head for detection results. To make sure that our model can quickly adapt to few-shot cases, we train it according to a two-phase training scheme like what transfer learning works do.

Based on RoI features, our model concentrates on object-level information and attach attention on it in order to alleviate the interference from objects of other classes. Through pre-training and parameter-sharing, our model is designed to be simple and lightweight to fit the circumstance where there are only a few annotated data. Simple as our model is, it is effective for few-shot object detection task due to a self-adaptively attention-added way. In addition, it is verified by the experiments on two VHR remote sensing datasets in different scales.

The main contributions of this paper are as follows.
（a）We propose a few-shot object detector based on the two-stage detector that concentrates on object-level relations in order to fit object-level data available in object detection task.
（b）We implement a Relation GRU unit to obtain the object-level relations and simultaneously add additional attention on the object features with the assistance of support images.
（c）Instead of meta-learning, we select transfer learning to alleviate difficulty in training procedure to fully leverage information from data in base classes.

## II. RELATED WORK

Object detection is a rather challenging task compared to classification task in computer vision. The methods of it can be divided into two categories with the help of deep CNNs: proposal-based and proposal-free. These two categories can also be commonly defined as two-stage and one-stage. R-CNN series detectors fall into the first category, all of which extract proposals at the beginning through CNNs and then aim to detect those proposals. These series detectors are derived from R-CNN [11], whose proposals are generated by selective search [12], and then are improved by SPP-Net [13] as well as Fast R-CNN [14]. Eventually, with the introduction of Region-Proposal-Network (RPN), Faster R-CNN achieves a rather better performance on both precision and speed compared to previous methods. In contrast, either YOLO series detectors [15]-[17] or SSD series [18] detectors can directly predict bounding boxes without generating region proposals through a single deep convolutional network by considering object detection task as regression task. Due to only one stage, these detectors target at lightweight network design and aim to high inference speed. In a word, they are fast but not as accurate as the first category.

Most works in object detection field rely on sufficient training data. However, there exist circumstances where training data are extremely rare. The research that focuses on learning from only a few training examples is called few-shot learning (FSL). To alleviate the problem of overfitting in FSL supervised learning, prior knowledge must be involved. Based on which aspect is enhanced with prior knowledge, existing FSL works can be categorized into the following three perspectives: "data", "model" and "algorithm". With prior knowledge, "data" [19] strands for the augmentation design of training dataset and "model" [20] means to constrain hypothesis space. As for "algorithm" [21], it aims to alter search strategy in hypothesis space by prior knowledge. Recently, great progress about classification task has been made in few-shot field. The popular solutions are mainly based on meta-learning [22]. Interestingly, [23] proposes that simple transfer-learning way with cosine-similarity based classifier can achieve a better performance compared to some methods based on meta-learning.

On the other hand, the challenging few-shot object detection task is far from fully researched. LSTD [24] introduces to combine part of SSD and part of Faster R-CNN as a whole detector to make full advantages of both. RepMet [25] is based on Distance Metric Learning (DML) and adds a DML-head to predictor head of detection network. Feature reweighting network [26] and Meta R-CNN [27] both target at training strategy of meta-learning style and attach attention on features in a direct way. All of the research on few-shot object detection task fall into the natural image scenery.

In remote sensing field, there are many works [28], [29] about object detection. However, almost all of them are based on the condition where there are sufficient training data. The concentration of those works is on how to make the detection networks adapt to remote sensing inputs. There also exist some works [30] focusing on few-shot classification in remote sensing field. However, as for the few-shot object detection, the study is rather few. [31] is a training-free design in remote sensing and target at sewage treatment plant and airport detections. Nevertheless, it is not able to be trained to be generalized to detect other objects and therefore its applications are severely restricted. In our work, to reduce the difficulty in training process, with the assistance of transfer learning, we aim at the challenging task in remote sensing field.

## III. PROPOSED METHOD

In many application scenarios, there exist data from some classes which are easy to get and thus in a large number. We call these classes base classes. On the contrary, the new classes of which the data are quite rare or cannot be easily acquired are novel classes. In these scenarios, we aim to obtain a few-shot detector that can be trained to detect novel object through fully leveraging knowledge from base classes. To this end, we design our model to explicitly take class-to-class relation information into consideration, which is different from existing methods that only make use of visual appearance clues. Apart from this, because of focusing RoIs (Region-of-Interest) instead of the whole images, the class-to-class relations are expressed in the object-object form and we attach importance to that additional information is fused in a self-adaptive way.

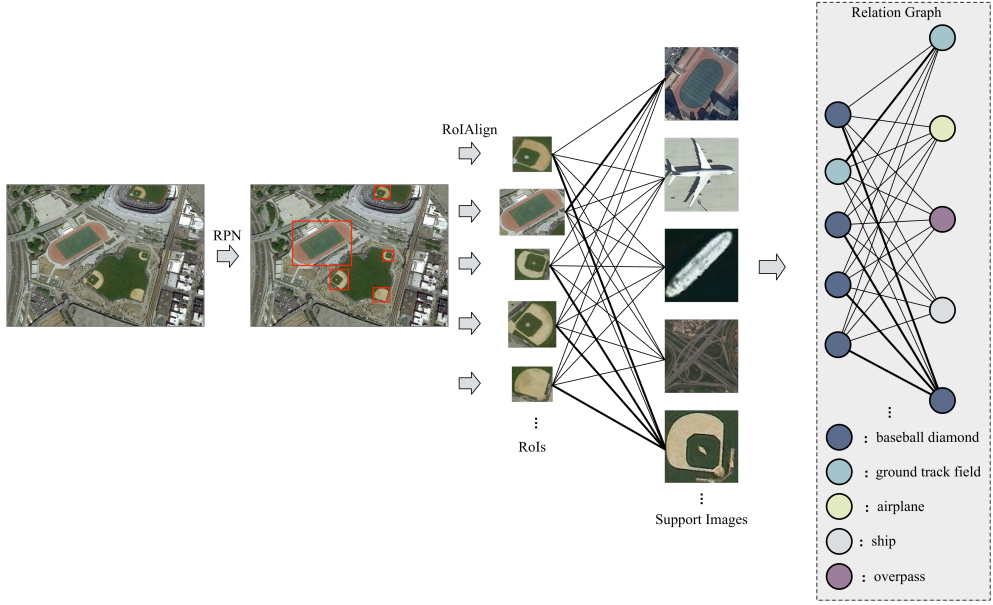

Fig. 1. Graphical Model. We define that the relation between RoIs and support images is formulated as a graphical model from a structure perspective.

*A. Model Definition*

Define In few-shot object detection task, a detector receives training data from base classes $C_{base}$ and novel classes $C_{novel}$. Therefore, the dataset can be divided into two groups:

$$D_{base} = \{(x_i^{base}, y_i^{base})\}_{i=1}^{n_1} \sim P_{base} \quad (1)$$

$$D_{novel} = \{(x_i^{novel}, y_i^{novel})\}_{i=1}^{n_2} \sim P_{novel} \quad (2)$$

In $D_{base}$, there are sufficient samples in each class. Whereas in $D_{novel}$, the number of samples in each class is quite small. Training only with samples from novel classes is infeasible because the detector will suffer severely overfitting due to large quantities of parameters to be trained in the huge detection network. Whereas with $D_{base} \cup D_{novel}$, the performance is still not satisfactory owing to the extreme data quantity imbalance between $D_{base}$ and $D_{novel}$.

Specifically, an image $x_i$ labeled with a structed annotation $y_i$ contains $n_i$ objects in different classes, sizes and positions. The few-shot learners are supposed to distinguish objects of novel classes from other objects and background. In other words, classes $c$ and predictions of positions $p$ should be provided by these learners. However, as most detection works do, modeling $h(x_i; \theta)$ is straightforward but not precise enough to overcome this difficulty. From another point of view, modeling $h(z_{i,j}; \theta)$ is necessary and explicit. Modeling with objects $\{z_{i,j}\}_{j=1}^{n_i}$ directly is able to alleviate the interference from other objects.

Taking Faster R-CNN as an example, the backbone of it takes an image $x_i$ as input and produces features: $f_i = D(x_i)$. After that, through RoIAlign on the image region proposals extracted by the region proposal network (RPN), RoIs are produced and the features $\{\hat{z}_{i,j}\}_{j=1}^{n_i}$ between objects from novel and base classes are detached naturally. To sum up, different from other detectors to model $h(x_i; \theta)$ on the whole image, it is of necessity to model $h(z_{i,j}; \theta)$ on RoI features $\{\hat{z}_{i,j}\}_{j=1}^{n_i}$ in order to reduce the impact from other objects.

As introduced previously, we define support set $S$ and query set $Q$. The images $x_{query}$ in query set $Q$ are exactly the images from the training or testing dataset. Whereas the images $x_{support}$ in support set $S$ are patches of objects cropped from the training dataset according to annotations. The images in support set are used to provide information of different classes in the attention form. In order to alleviate distortion, we crop them by making up the short edge to be equal to the long edge and then resize them into $224 \times 224$ for simplicity:

$$\begin{cases} w^i = a_2^i - a_1^i \\ h^i = b_2^i - b_1^i \end{cases} \quad (3)$$

$$\begin{cases} b_1^i = max\left(b_1^i - \frac{w^i - h^i}{2}, 0\right) \\ b_2^i = min\left(b_2^i + \frac{w^i - h^i}{2}, H\right) \end{cases}, if\ w^i \geq h^i \quad (4)$$

$$\begin{cases} a_1^i = max\left(a_1^i - \frac{h^i - w^i}{2}, 0\right) \\ a_2^i = min\left(a_2^i + \frac{h^i - w^i}{2}, W\right) \end{cases}, if\ w^i < h^i \quad (5)$$

where $(a_1^i, b_1^i)$ denotes the positions of the left-top point of an annotation and $(a_2^i, b_2^i)$ denotes the right-bottom point. As for $w^i$ and $h^i$, they are the width and height of the annotation respectively. $W$ and $H$ denotes the width and height of the image. The positions $(a_1^i, b_1^i)$ and $(a_2^i, b_2^i)$ on which support images are to be cropped based are calculated according to (4) and (5). Furthermore, this design is able to mix some background information around the objects to assist detection as well.

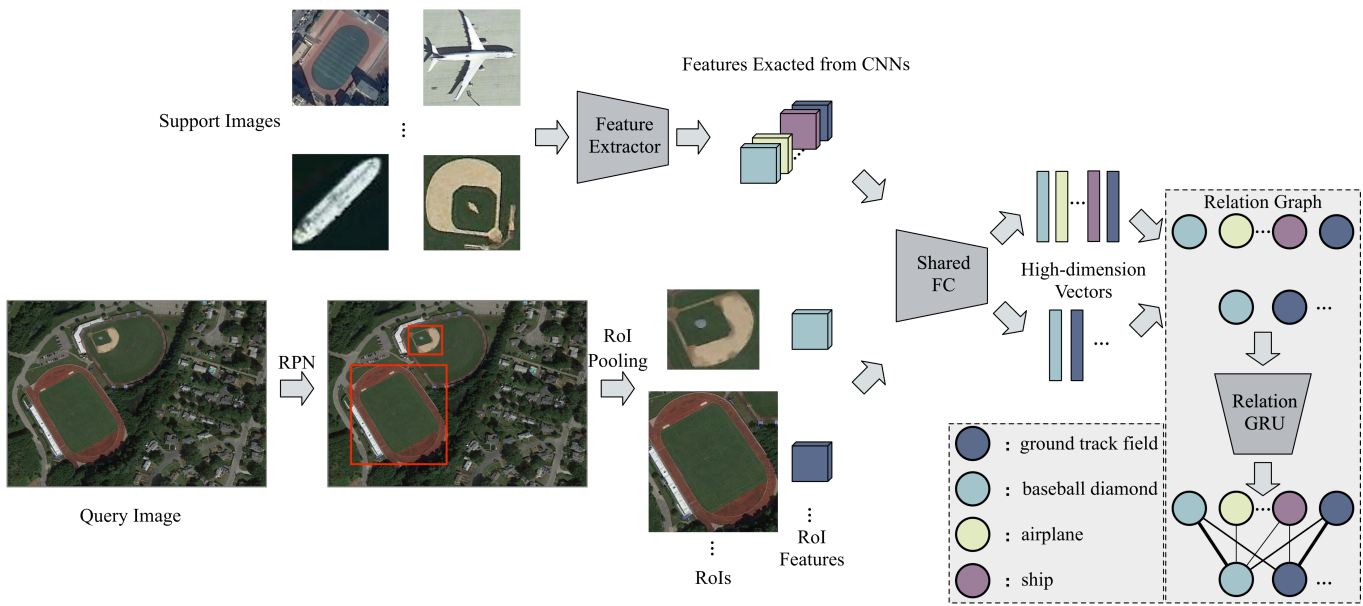

Fig. 2. The architecture of our proposed few-shot object detector based on Faster R-CNN. In the top line, support images are extracted into features and then processed into high-dimension vectors. Similarly, the query image is processed and then fused with information from support images through Relation GRU.

As shown in Fig. 1, we define a graphical relation model $G = (V, E)$ to model the relation between RoI features and features of support images. The left nodes $v_{left} \in V$ represent features of region proposals whereas the right nodes $v_{right} \in V$ represent features of support images. The edges $e \in E$ between left nodes $v_{left}$ and right nodes $v_{right}$ are their corresponding relation. Naturally, detecting some classes, such as ground track field, can be assisted with the information from the classes themselves or some other classes such as baseball diamond and basketball court. However, some information from other classes, such as airplane, can be detrimental for detecting them. As a consequence, it is of great significance to model the relation between classes or even between objects due to diversity of colors and shapes in one single class. Under this circumstance, graph can be an explicit and natural way to model the relation.

### B. Self-Adaptive Attention Network (SAAN)

Our proposed few-shot detector is based on Faster R-CNN, which is a classic but commonly-used two-stage object detector. The reason why we choose a two-stage object detector is that the process of how it detects objects is explicit and clear compared with one-stage object detectors. In addition, we choose Faster R-CNN because it is classic and more importantly it can separate objects in a natural way. The architecture of our proposed few-shot object detector is in Fig. 2. The bottom line to process a query image is as what Faster R-CNN does to obtain features of images in the beginning through RCNN-base composed of CNNs. Secondly, with the assistance of RPN and RoIAlign method, RoIs are produced and then are fed into RoI-head that is made up of some fully-connected layers. Ultimately, before predicted by predict-head, these high-dimensional vectors that contains information of the possible objects are fused with the information of the support images.

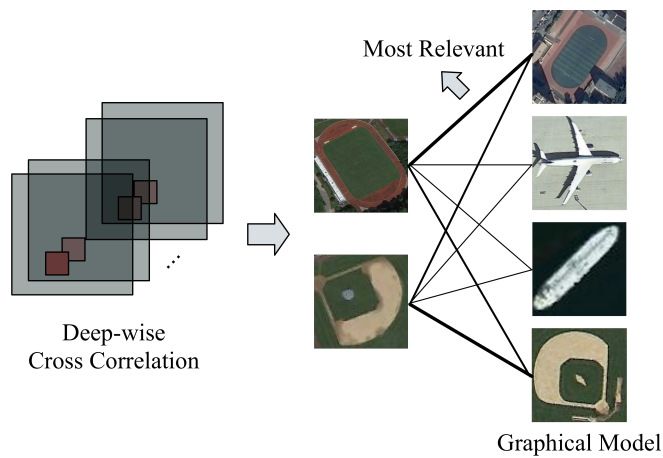

Fig. 3. Compared to deep-wise cross correlation, the relation between information is modelled by graph.

On the other hand, the top line shows how to process support images that contains information of objects from base and novel classes. The feature extractor extracts features of support images in a batch-to-batch form. The same as what features of a query images are processed, these features are fed into some fully-connected layers. To alleviate overfitting, the parameters of the feature extractor and the fully-connected layers to be trained here are shared with the RCNN-base and RoI-head. Additionally, the high-dimensional vectors of these supports images are almost obtained by the same way of what the ones of the query image are processed. As a result, they can be modelled to be graphical possibly.

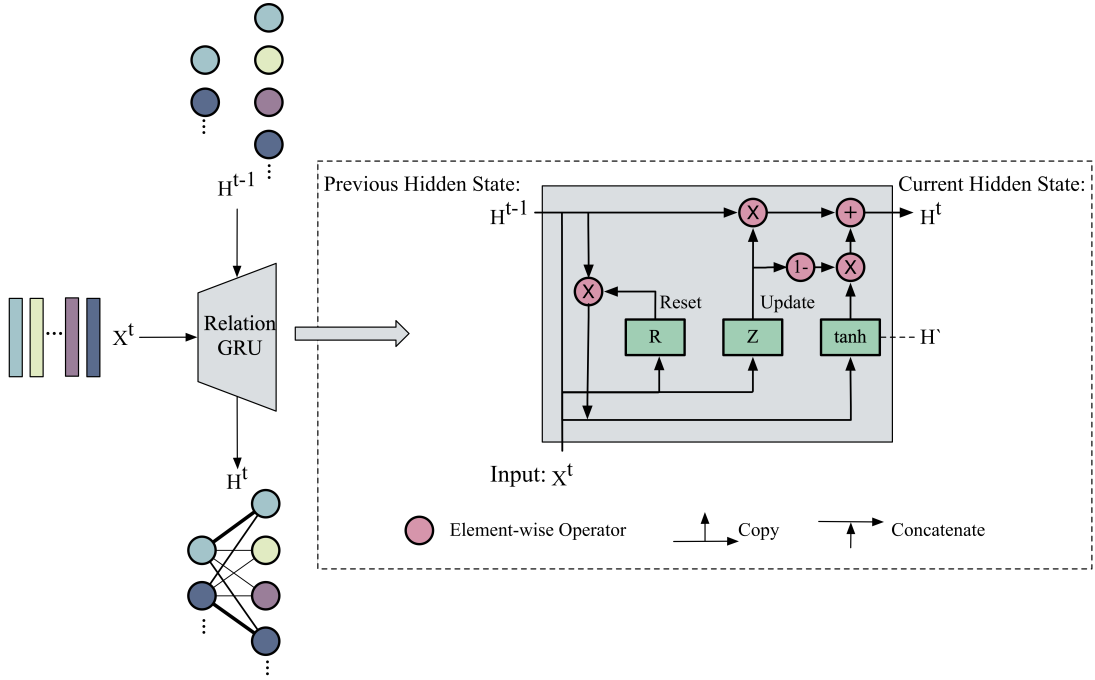

Fig. 4. The architecture of our Relation GRU. It takes information as inputs and updates the graphical relation.

As talked before, the vectors of a query image contain information of probable objects. Nevertheless, the information of the objects from novel classes is not sufficient enough to be predicted. Therefore, it is supposed to be fused with the information from support images as what an attention does. Specifically, in Fig. 3, the attention that is added here by relation GRU (Gate Recurrent Unit [32]) is in a self-adaptive form, which is different from the commonly-used deep-wise cross correlation. The deep-wise cross correlation is an operation which calculates the similarity between two units directly by pointwise multiplication. It is simple but not precise enough to consider some cases where the information added can be detrimental.

On the other hand, as modelled to be graphical, each left node receives messages delivered from the right nodes. Because there are multiple messages from every single right node, it is necessary to aggregate them with the information or what we call "memory" of the node itself and then fuse these messages into meaningful representations. Obviously, there exist structures called RNNs (Recurrent Neural Network) that is able to achieve this memory-selection function. In detail, there are recurrent connections in RNNs, which can convert a memory of previous inputs into internal state and thereby influence the output. Particularly, compared with simple RNN and complex LSTM (Long-Short Term Memory [33]), GRU that is popular in NLP (Natural Language Processing), is a promising choice because it performs similarly to LSTM but is computationally cheaper. In short, GRU is an effective and lightweight unit to obtain "memory" function.

The internal architecture of GRU is in Fig. 4. In the beginning, the reset gate $R$ is computed by

$$R = \sigma(W^r con[X^t, H^{t-1}]) \quad (6)$$

where $\sigma$ is the sigmoid activate function and $con[,]$ denotes to concatenate two tensors. $W^r$ is a weight matrix to be learned and $H^{t-1}$ is previous hidden state which has the same dimension compared with input $X^t$. In a similar way, the update gate $Z$ is computed by

$$Z = \sigma(W^z con[X^t, H^{t-1}]) \quad (7)$$

Secondly, the state is reset by

$$H' = \tanh(WX^t + U(R \odot H^{t-1})) \quad (8)$$

where $W$ and $U$ are weight matrices that are to be learned. $\odot$ denotes the element-wise multiplication. Eventually, the current hidden state is computed by

$$H^t = ZH^{t-1} + (1-Z)H' \quad (9)$$

To sum up, the GRU is an effective unit that can remember long-term memory. Through reset gate $R$, this unit is able to drop the information of inputs that is irrelevant with the hidden state. Whereas through update gate $Z$, it controls how much information to retain in current hidden state. In other words, this unit with learnable weight matrices performs like an aggregation to add self-adaptive attention. Specifically, we utilize features from support images as input and features from RoIs as initial hidden state. In this way, additional parameters though this unit brings in, the inputs of it are a great number of objects cropped from support images. As a consequence, the overfitting problem is alleviated.

## C. Training Scheme

Different from most few-shot works that train models in meta-learning form, we adapt a two-phase training scheme in our work like transfer learning task to ensure that our model can obtain a generalized performance in few-shot settings. As shown in Fig. 5, our training phase is divided into two parts: base training and novel fine-tuning.

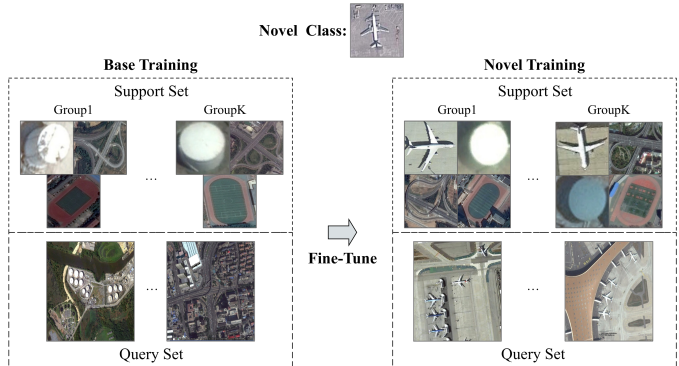

Fig. 5. Our two-phase training scheme: we train our model only with base classes firstly and then fine-tune it with additional novel classes.

In base training phase, only base classes with sufficient labeled data are available. As talked before, the images in query set are directly from our dataset. The ones in support set are cropped from query set according to the annotation. In the second phase, the novel fine-tuning phase, the model is trained on not only base but also novel classes. The training data are the same like previous phase except that there are images from novel classes. Because there are only $k$ annotations for each novel class, we also choose $k$ annotations for each base class to strike balance between them and we will discuss this setting in Experiments. Moreover, the procedure of the second phase is the same as the first one.

## IV. EXPERIMENTS

### A. Experiment Datasets and Baselines

We evaluated the performance of our proposed networks on the dataset RSOD [34], which is a 4-class geospatial dataset for object detection. These 4 classes in RSOD are aircraft, oil tank, overpass, and playground. Among these 4 classes, 1 class is randomly selected to become the novel class, whereas the rest 3 classes are selected to be base ones. Moreover, we evaluate the performance with 4 different base/novel splits for objectivity. In first phase, as for each base class, we randomly select 80% of the images to train our model then the rest ones for testing to verify the performance of our proposed approach and the backbone is the one already pre-trained with ImageNet [35]. In novel fine-tuning phase, we only choose quite a small set of training images to make sure that there are only $k$ annotated bounding boxes in each class, where $k$ equals 1, 2, 3, 5, and 10 respectively.

To further verify the generalization of our model, we also evaluated the performance on the dataset NWPU VHR-10 [36] with 10 classes that is more difficult than RSOD. The 10 classes in RSOD are airplane, ship, storage tank, baseball diamond, tennis court, basketball court, ground track field, harbor, bridge, and vehicle. Particularly, we exclude storage tank and harbor from our evaluation on this dataset because the number of objects in one image in these classes is much larger than 10 and thus this cannot satisfy our experimental settings. Except for that, the base/novel splits and other dataset settings are the same as RSOD. Eventually, we conduct ablation study based on RSOD dataset as well. Specifically, we analyze the proportion of the number of objects in each base class $n_{base}$ compared with the number of objects in each novel class $n_{novel}$ utilized in the second phase.

As for baselines, our model is compared with three ones. The first baseline is to train the faster R-CNN with images from the base and novel classes together. There are abundant annotations in base classes whereas novel classes fit the few-shot condition. We define it as FRCN-joint. The second one is trained with two training phases like ours. In detail, train the original pre-trained Faster R-CNN model the same as ours with only base classes in base training phase. Then in novel fine-tuning phase, the model is fine-tuned with base as well as novel classes. We define it as FRCN-ft. The rest one is meta-R-CNN which is a few-shot object detection model for natural images based on Faster R-CNN as well. With the help of these three baselines, the function of self-adaptive attention is verified and the performance of our SAAN is fully evaluated.

### B. Performance Analysis

The results of our experiment evaluated with AP of novel class on RSOD dataset are in TABLE I. First, in most conditions, our method outperforms the baselines which indicates the effectiveness and stability of our model. In general, compared with baselines, our model is almost either state-of-the-art or the second best. Moreover, the results of experiments on different splits of dataset show the generalized performance of our model when meeting different few-shot cases. Second, the performance of baseline 2 is better than the one of baseline 1 in most cases, which indicates that training on base classes in the beginning and then fine-tuning is an effective way to alleviate the imbalance between the number of base and novel classes. Nevertheless, the performance of FRCN is still poor to detect novel classes and it indicates that the difficulty cannot be totally overcome with transfer training strategy. As for meta R-CNN, our model outperforms it in the majority of the cases. In addition, in some setups such as playground for novel class, our mothed exceed meta R-CNN significantly. Although there exist cases where our method cannot outperform, the performances are nearly the same. Eventually, when taking playground as novel class, simply with only 10 objects to train our model, the precision of detection is up to almost 100%.

Generally speaking, when we focus on different objects, the difficulty of detecting them is distinct from class to class. For instance, the number of 10 or 10-shot of annotations of aircraft is completely not enough to detect it. However, as for oil tank or playground in the same experimental condition, it is much more adequate. As a consequence, when it comes to few-shot learning research, the number of the objects that we define as few, may about to be different with the changes of the object we study.

TABLE I
EXPERIMENTAL RESULTS OF FEW-SHOT OBJECT DETECTION (AP OF NOVEL CLASS) ON RSOD DATASET. PARTICULARLY, RED AND BLUE INDICATE STATE-OF-THE-ART (SOTA) AND THE SECOND BEST.

| Novel Class | aircraft | | | | | oil tank | | | | |
|---|---|---|---|---|---|---|---|---|---|---|
| Method/Shot | 1 | 2 | 3 | 5 | 10 | 1 | 2 | 3 | 5 | 10 |
| FRCN-joint | 0.19 | 9.09 | 9.09 | 9.91 | 17.04 | 9.09 | 9.39 | 11.23 | 17.86 | 59.84 |
| FRCN-ft | 2.27 | 6.26 | 10.76 | 16.05 | 31.34 | 13.09 | 14.66 | 48.04 | 57.94 | 72.54 |
| Meta R-CNN | 9.09 | 9.09 | 18.40 | 35.03 | 45.13 | 15.09 | 19.54 | 58.45 | 60.14 | 81.69 |
| Ours | 10.80 | 12.34 | 23.20 | 34.38 | 45.09 | 26.46 | 28.26 | 53.43 | 67.02 | 81.42 |

| Novel Class | overpass | | | | | playground | | | | |
|---|---|---|---|---|---|---|---|---|---|---|
| Method/Shot | 1 | 2 | 3 | 5 | 10 | 1 | 2 | 3 | 5 | 10 |
| FRCN-joint | 1.32 | 3.79 | 10.98 | 25.52 | 32.88 | 9.35 | 19.15 | 37.96 | 31.73 | 76.12 |
| FRCN-ft | 2.31 | 5.77 | 12.73 | 16.28 | 31.64 | 28.84 | 34.95 | 52.18 | 75.99 | 90.58 |
| Meta R-CNN | 2.68 | 4.67 | 11.36 | 32.63 | 42.66 | 30.6 | 45.6 | 62.7 | 81.33 | 88.11 |
| Ours | 4.71 | 9.01 | 10.42 | 34.09 | 44.13 | 38.34 | 52.48 | 77.72 | 90.22 | 96.15 |

TABLE II
EXPERIMENTAL RESULTS OF FEW-SHOT OBJECT DETECTION (AP OF NOVEL CLASS) ON NWPU VHR-10 DATASET EXCEPT FOR STORAGE TANK AND HARBOR CLASSES. PARTICULARLY, RED AND BLUE INDICATE SOTA AND THE SECOND BEST.

| Novel Class | airplane | | | | | ship | | | | |
|---|---|---|---|---|---|---|---|---|---|---|
| Method/Shot | 1 | 2 | 3 | 5 | 10 | 1 | 2 | 3 | 5 | 10 |
| FRCN-joint | 4.55 | 9.09 | 13.13 | 17.82 | 18.06 | 31.07 | 39.21 | 42.66 | 72.01 | 77.56 |
| FRCN-ft | 9.09 | 14.27 | 13.98 | 18.65 | 32.99 | 21.55 | 42.83 | 55.51 | 73.58 | 78.94 |
| Meta R-CNN | 9.09 | 17.82 | 23.63 | 29.25 | 40.91 | 25.53 | 52.73 | 64.90 | 73.09 | 79.62 |
| Ours | 9.09 | 15.72 | 25.71 | 34.68 | 38.78 | 28.77 | 52.78 | 57.95 | 78.05 | 81.14 |

| Novel Class | baseball diamond | | | | | tennis court | | | | |
|---|---|---|---|---|---|---|---|---|---|---|
| Method/Shot | 1 | 2 | 3 | 5 | 10 | 1 | 2 | 3 | 5 | 10 |
| FRCN-joint | 22.00 | 59.03 | 56.80 | 76.84 | 87.39 | 9.09 | 15.15 | 17.75 | 42.62 | 44.22 |
| FRCN-ft | 15.87 | 62.33 | 78.17 | 79.93 | 88.56 | 14.74 | 17.73 | 25.69 | 41.10 | 53.08 |
| Meta R-CNN | 28.60 | 60.41 | 78.28 | 88.18 | 74.09 | 16.74 | 17.02 | 26.02 | 43.16 | 44.37 |
| Ours | 32.60 | 63.41 | 80.26 | 81.20 | 89.75 | 18.32 | 23.73 | 29.45 | 47.68 | 54.11 |

| Novel Class | basketball court | | | | | ground track field | | | | |
|---|---|---|---|---|---|---|---|---|---|---|
| Method/Shot | 1 | 2 | 3 | 5 | 10 | 1 | 2 | 3 | 5 | 10 |
| FRCN-joint | 6.92 | 7.41 | 21.42 | 40.09 | 50.86 | 17.20 | 21.94 | 26.09 | 66.69 | 81.88 |
| FRCN-ft | 13.80 | 20.73 | 28.19 | 45.48 | 71.57 | 23.31 | 39.07 | 46.91 | 71.15 | 88.81 |
| Meta R-CNN | 18.77 | 20.60 | 22.35 | 45.53 | 64.53 | 32.20 | 38.58 | 55.94 | 55.65 | 52.08 |
| Ours | 21.33 | 37.74 | 42.43 | 54.78 | 80.16 | 31.29 | 42.24 | 52.12 | 78.88 | 89.57 |

| Novel Class | bridge | | | | | vehicle | | | | |
|---|---|---|---|---|---|---|---|---|---|---|
| Method/Shot | 1 | 2 | 3 | 5 | 10 | 1 | 2 | 3 | 5 | 10 |
| FRCN-joint | 0.61 | 9.09 | 13.34 | 16.04 | 16.35 | 9.09 | 9.09 | 15.55 | 24.33 | 27.04 |
| FRCN-ft | 0.15 | 4.89 | 11.87 | 14.37 | 23.72 | 9.09 | 12.02 | 20.16 | 27.73 | 34.23 |
| Meta R-CNN | 0.10 | 9.09 | 9.28 | 9.09 | 23.90 | 9.09 | 9.88 | 14.51 | 18.26 | 24.58 |
| Ours | 3.08 | 9.31 | 16.94 | 18.68 | 43.94 | 9.37 | 13.15 | 22.77 | 29.60 | 36.24 |

In TABLE II, we demonstrate the comparison of the detection results on NWPU VHR-10 dataset. More intuitively, we illustrate the results in the form of histograms in Fig. 6. Compared to RSOD dataset, there are more classes but less images or objects in each class in NWPU VHR-10. In this setting, the dataset is more challenging than RSOD. The same like the results on RSOD, our model outperforms FRCN baselines in almost every case. On the other hand, when the shots of novel classes increase, the AP of the detection improves. These two improvements indicate that the information about a novel class comes from two parts: one is from the samples of the class itself and the other one is from the samples of other classes. When the information comes from other classes, the way to utilize it better is to fine-tune rather than training together.

As for meta R-CNN, its performance is unsatisfactory and it performs even poorly than FRCN in some cases. This is probably because the dataset is much more complex, and it is inferred that the way to implement attention directly may lead to harmful impact on detection. Though the messages from a variety of classes is confusing, our method is able to fuse them into meaningful representations by our SAAN with graphical model. In short, the more exact information we extract, the better results of detection will show. As a consequence, the results demonstrate that our proposed method is able to obtain information from other classes in a more effective way when conducting fine-tuning method.

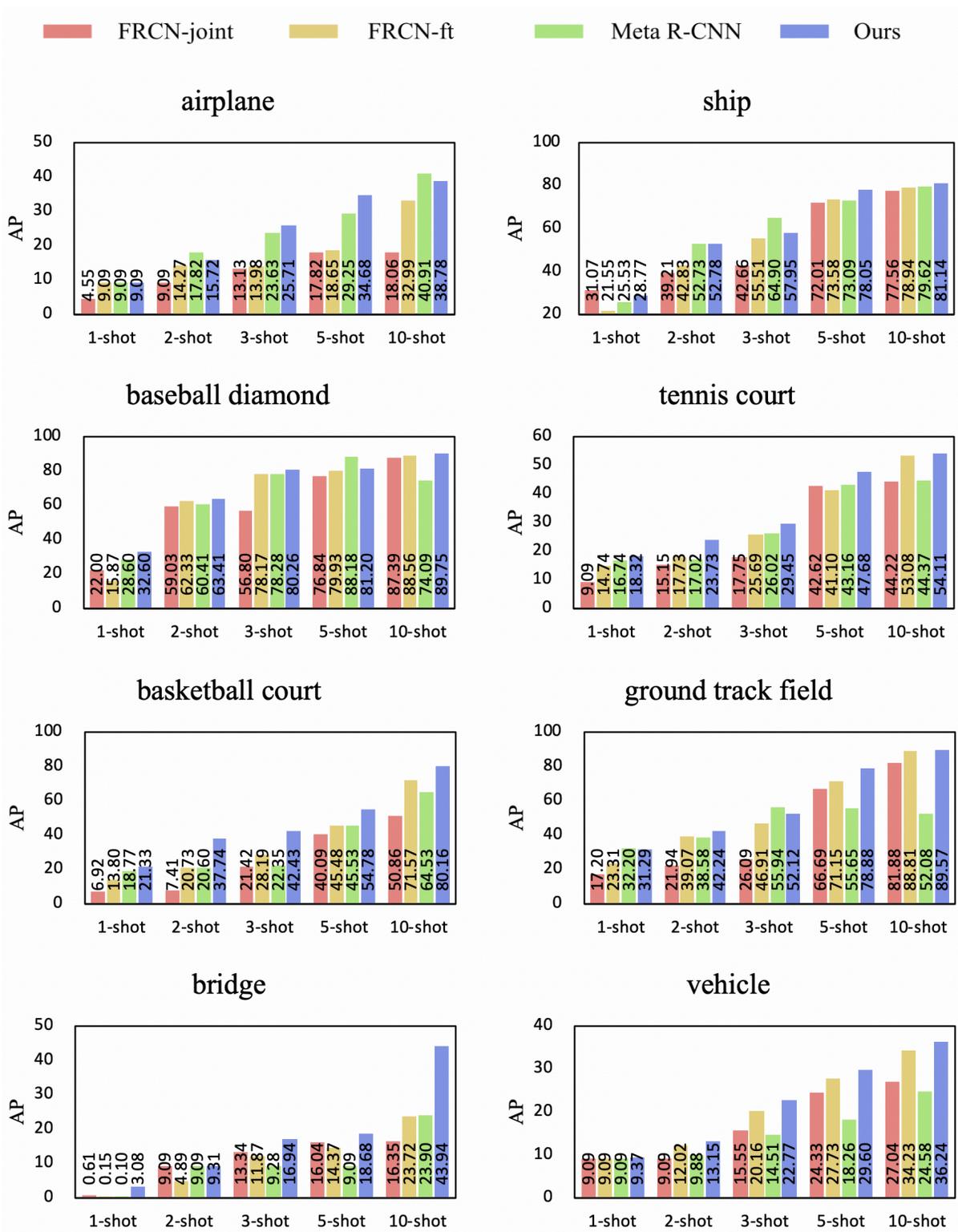

Fig. 6. AP of novel objects with histograms on NWPU VHR-10.

TABLE III
EXPERIMENTAL RESULTS OF DIFFERENT PROPORTION BETWEEN BASE AND NOVEL OBJECTS UTILIZED IN THE SECOND TRAINING PHASE ON RSOD DATASET.

| Shot | Propotion(Base : Novel) | Novel Class | Base Class | | |
|---|---|---|---|---|---|
| | | aircraft | oil tank | overpass | playground |
| 1 | 0 (only novel) | **14.59** | 73.64 | 7.03 | 71.81 |
| | 1 : 1 (selected) | 10.8 | 90.26 | 46.47 | 90.57 |
| | 2 : 1 | 9.44 | 90.62 | 65.69 | 90.91 |
| | 3 : 1 | 9.48 | 90.7 | 68.24 | 90.91 |
| | 5 : 1 | 9.09 | 90.73 | 56.65 | 1 |
| | ∞(all base objects used) | 9.09 | **97.48** | **88.77** | 1 |
| 2 | 0 (only novel) | 15.12 | 53.26 | 4.04 | 78.73 |
| | 1 : 1 (selected) | 12.34 | 90.57 | 67.66 | 90.96 |
| | 2 : 1 | **16.37** | 90.59 | 65.65 | 99.39 |
| | 3 : 1 | 11.34 | 90.66 | 39.42 | 98.79 |
| | 5 : 1 | 11.35 | 90.67 | 55.86 | 1 |
| | ∞(all base objects used) | 10.44 | **90.83** | **81.56** | 1 |
| 3 | 0 (only novel) | 18.98 | 61.56 | 1.71 | 58.21 |
| | 1 : 1 (selected) | 23.2 | 90.45 | 64.73 | 90.91 |
| | 2 : 1 | 20.14 | 90.6 | 33.03 | 90.91 |
| | 3 : 1 | 23.23 | 90.59 | 50.75 | 1 |
| | 5 : 1 | **24.6** | 90.59 | 59.81 | 99.69 |
| | ∞(all base objects used) | 18.17 | **90.76** | **81.44** | 1 |
| 5 | 0 (only novel) | 23.66 | 36.18 | 1.19 | 69.72 |
| | 1 : 1 (selected) | **34.38** | 90.59 | 64.46 | 1 |
| | 2 : 1 | 34.27 | 90.48 | 53.94 | 1 |
| | 3 : 1 | 30.47 | 90.67 | 63.68 | 1 |
| | 5 : 1 | 29.54 | 90.78 | 68.29 | 99.12 |
| | ∞(all base objects used) | 26.53 | **90.83** | **81.52** | 1 |
| 10 | 0 (only novel) | 32.31 | 51.68 | 10.63 | 79.62 |
| | 1 : 1 (selected) | **45.09** | 90.65 | 47 | 99.69 |
| | 2 : 1 | 42.12 | 90.7 | 64.64 | 99.39 |
| | 3 : 1 | 42.63 | **90.81** | 71.79 | 90.91 |
| | 5 : 1 | 42.55 | 90.72 | 63.45 | 90.91 |
| | ∞(all base objects used) | 42.94 | **90.81** | **89.91** | 1 |

### C. Ablation Study and Visualization

We demonstrate our results on ablation study in TABLE III. In the phase when we fine-tune our model already trained by important to analyze the proportion between them. From the perspective of experimental results, fine-tuning model only with novel objects do not work well neither on novel nor on base classes. It may be ascribed to that the information from novel class itself is not adequate for detection. As for base classes, the model suffers catastrophic forgetting [37] under this circumstance with the training scheme like transfer learning. On the other hand, as this proportion increases, the performances on base classes are getting better. Particularly, when we utilize all base objects in the second phase, the model achieves the best performance on base classes. Nevertheless, the novel objects are drowned in the base objects and therefore the performance on novel class is poor. Furthermore, it is computational-consuming and thus a waste of time to fine-tune with so much base objects especially when we concentrate on novel objects. In conclusion, we choose to strike balance between the number of base and novel objects in order to compromise between effectiveness and efficiency.

Ultimately, we provide qualitative visualizations of the detected novel objects on NWPU VHR-10 in Fig. 7. The top row shows the detection results through FRCN-ft method and the middle row illustrates the results through meta R-CNN method. Whereas detection results through our SAAN method are illustrated in the bottom row. The visualizations are based on treating bridge as novel class whereas airplane, baseball diamond and other classes as base classes. The left and middle columns show the results of detecting these two base classes and the results of detecting novel class is shown in the right column.

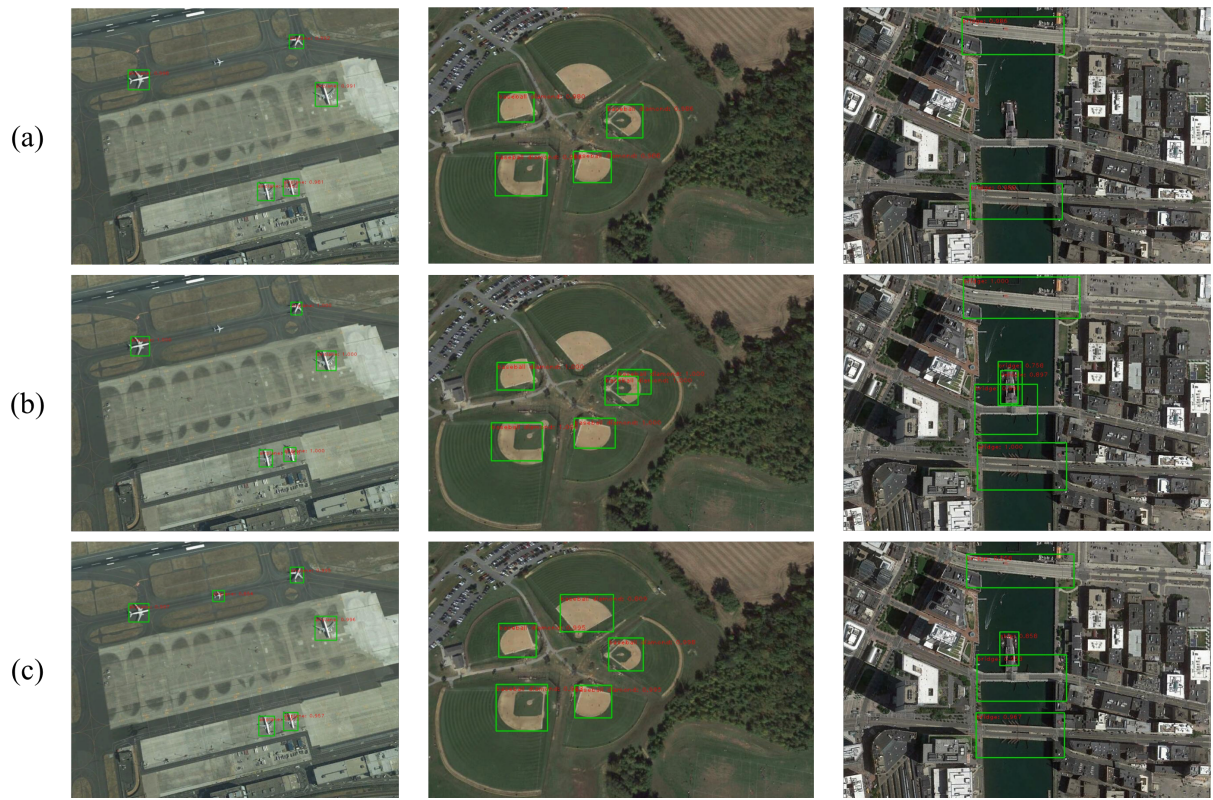

Fig. 7. Comparison of the detection results for the FRCN-ft, meta R-CNN and our SAAN methods. (a) Detection results of the FRCN-ft method. (b) Detection results of the meta R-CNN method. (c) Detection results of our SAAN method.

When concentrating on novel class, there are missed detection or error detection of base classes in FRCN-ft and meta R-CNN method. As for novel class, FRCN-ft miss some objects ascribed to lacking of additional information. On the other hand, as for meta R-CNN, though it does not miss novel objects (bridges), it fails to detect the base object (ship) beside the bridge. Because there is no filtration of information from other classes, the ship is mistaken as bridge or even vehicle (another base class). As demonstrated in the bottom row, our method strikes a balance when detecting base and novel classes. With the assistance of SAAN by graphical model, the information is fused in a self-adaptive way. As a consequence, the detection results are not interfered by each other. In short, the qualitative visualizations demonstrate that our model achieves less missed and error detection and it provides more accurate positions of objects.

## V. CONCLUSION

In this paper, we proposed a few-shot object detection model that is able to detect novel object with only a few annotated data in remote sensing images. Our proposed model is based on two-stage detectors. For simplicity, we implement Faster R-CNN as detector, which is a commonly-used object detection framework and train it in a transfer learning scheme. With the help of the Self-Adaptive Attention Network (SAAN), our graphical model is able to fully leverage object-level knowledge from base objects in a self-adaptive way and selectively adapt to novel classes. Experimental results based on the RSOD and NWPU VHR-10 dataset verify the effectiveness of our model when faced with few-shot cases in remote sensing. Few-shot object detection in remote sensing is a challenging problem and we will further explore how to improve its performance for more complex scenes.